\documentclass[lettersize,journal]{IEEEtran}
\usepackage{amsmath,amsfonts}
\usepackage{algorithmic}
\usepackage{algorithm}
\usepackage{array}
\usepackage[caption=false,font=normalsize,labelfont=sf,textfont=sf]{subfig}
\usepackage{textcomp}
\usepackage{stfloats}
\usepackage{url}
\usepackage{verbatim}
\usepackage{graphicx}
\usepackage{cite}
\usepackage{cuted}
\usepackage{enumitem}
\usepackage{float}
\usepackage{multirow}
\usepackage{xcolor}
\usepackage{booktabs}
\usepackage{makecell}
\usepackage[colorlinks=true, linkcolor=black, filecolor=black, urlcolor=black, citecolor=black]{hyperref}
\begin{document}

\title{\textbf{pyrtklib}: An open-source package for tightly coupled deep learning and GNSS integration for positioning in urban canyons}

\author{Runzhi Hu, Penghui Xu, Yihan Zhong, and Weisong Wen*
        % <-this % stops a space
\thanks{All authors are with the Hong Kong Polytechnic University. (The corresponding author to provide e-mail: welson.wen@polyu.edu.hk)}% <-this % stops a space
}

\markboth{}%
{}

\IEEEpubid{}

\maketitle

\begin{abstract}
Artificial intelligence (AI) is revolutionizing numerous fields, with increasing applications in Global Navigation Satellite Systems (GNSS) positioning algorithms in intelligent transportation systems (ITS) via deep learning. However, a significant technological disparity exists as traditional GNSS algorithms are often developed in Fortran or C, contrasting with the Python-based implementation prevalent in deep learning tools. To address this discrepancy, this paper introduces pyrtklib, a Python binding for the widely utilized open-source GNSS tool, RTKLIB. This binding makes all RTKLIB functionalities accessible in Python, facilitating seamless integration. Moreover, we present a deep learning subsystem under \textbf{pyrtklib}, which is a novel deep learning framework that leverages pyrtklib to accurately predict weights and biases within the GNSS positioning process. The use of pyrtklib enables developers to easily and quickly prototype and implement deep learning-aided GNSS algorithms, showcasing its potential to enhance positioning accuracy significantly.
\end{abstract}

\begin{IEEEkeywords}
Artificial intelligence, Deep learning, GNSS, RTKLIB
\end{IEEEkeywords}

\section{Introduction}

\IEEEPARstart{W}{ith} the rapid growth of computing speed and power, artificial intelligence (AI), epitomized by deep learning (DL), is now practical for everyday use. Due to its excellent non-linear fitting capability, deep learning has proven effective in numerous fields, including computer vision (CV) and natural language process (NLP). In modern intelligent transportation systems (ITS), deep learning also demonstrates potential in areas such as traffic control\cite{chu2021traffic}, autonomous driving (AD)\cite{zhu2021survey}, and human behavior analyze\cite{li2022new}. Global positioning, commonly known as global navigation satellite system (GNSS) positioning, is crucial for perception and decision-making within ITS\cite{zhang20203d,heng2014gnss,chen2020estimate}. Consequently, there is a pressing need within the ITS community to integrate deep learning techniques into global positioning strategies.

Currently, GNSS positioning accuracy can achieve centimeter-level precision under open skies. However, in urban canyons, the performance dramatically declines as GNSS signals are diffracted, reflected, and even obstructed by high-rise buildings\cite{ng2021urban,zhang20203d}. To mitigate the adverse effects of non-line-of-sight (NLOS) and multipath interference, two strategies are implemented: directly correcting measurements to improve accuracy and downweighing measurements from low-quality signals in the weight least squares (WLS) solution process\cite{groves2015principles}. Traditional methods use physical models and empirical formulas to model biases and weights, providing high interpretability but often struggling in complex and dynamically changing environments. In contrast, data-driven deep learning approaches can potentially model biases and weights more effectively, provided that the training data is of high quality.

Deep learning frameworks and applications have overwhelmingly adopted Python due to its simplicity, flexibility, and the vast ecosystem of libraries and tools available, such as TensorFlow \cite{abadi2016tensorflow} and PyTorch \cite{paszke2019pytorch}. This accessibility and ease of use facilitate rapid prototyping and deployment of complex models, making Python the language of choice for most new developments in deep learning and AI. Meanwhile, GNSS software historically leans on programming languages like Fortran and C \cite{dach2015bernese,takasu2009development,pany2019multi}. Though these languages offer high performance and control over hardware interaction, which are critical for the real-time processing demands and precision required in satellite navigation, these languages are not equipped with good compatibility with Python. This divergence in technological stacks presents a significant challenge for integrating cutting-edge AI methodologies, like deep learning, directly into traditional GNSS software systems. Bridging this gap requires either extending these systems to interface with Python-based tools or developing new capabilities within the GNSS software to support advanced machine learning techniques directly in C or Fortran, both of which entail substantial development and potential refactoring of existing codebases.

\IEEEpubidadjcol

To fill this gap, we make the Python binding, named \textbf{pyrtklib}, for the most popular open-source GNSS library, RTKLIB\cite{takasu2009development}. \textbf{pyrtklib} provides access to the full functionalities of RTKLIB, combining the speed of C with the convenience of Python. Additionally, we introduce a deep learning subsystem within \textbf{pyrtklib} for predicting weights and biases, which has been validated on our datasets. The following contributions of this paper are presented:

\begin{enumerate}

\item This paper developes a network to predict pseudorange biases and integrates these predictions into the correction of GNSS pseudorange measurements, tightly coupling them within the least squares process.

\item This paper has designed a network specifically to predict weights for each measurement, which are then utilized in the weighted least squares process.

\item This paper presents a network that simultaneously predicts both weights and biases for each measurement, applying these predictions in the weighted least squares process to improve accuracy.

\item This paper has open-sourced a Python package named \textbf{pyrtklib}, a Python binding for RTKLIB. Using meta-programming techniques, we automatically translate RTKLIB’s header files into Python binding code via pycparser\cite{Bendersky2022} and pybind11\cite{Jakob2016}, maintaining the integrity of RTKLIB’s constants and functions. The package can be accessed at \href{https://github.com/IPNL-POLYU/pyrtklib}{https://github.com/IPNL-POLYU/pyrtklib}.

\item Building on \textbf{pyrtklib}, this paper proposes a subsystem that integrates deep learning into the GNSS positioning process. This innovative framework is engineered for training and predicting weights and biases within the least squares solving process, available at \href{https://github.com/ebhrz/TDL-GNSS}{https://github.com/ebhrz/TDL-GNSS}.

\end{enumerate}

\section{Related Works}
\subsection{GNSS tools}

There are several existing GNSS tools as shown in Table \ref{GNSS_tools}. Bernese GNSS\cite{dach2015bernese} and MuSNAT\cite{pany2019multi} are two popular commercial GNSS software, however, their closed-source nature limits their usability for algorithm development. NavSU, MAAST, and goGPS\cite{herrera2016gogps} are open-source software programs written in Matlab. Despite their open-source status, a paid Matlab license is still required for their use. Laika and gnss\_lib\_py both are Python libraries, but they only provide basic GNSS functions and lack the support for real-time kinematic (RTK) and precise point positioning (PPP). RTKLIB is a comprehensive open-source GNSS tool that enjoys widespread use not only within the GNSS community but also in the robotics sector\cite{yin2021m2dgr,cao2022gvins,li2022p,kilic2021slip}. However, its C-based architecture poses integration challenges with Python. Additionally, while the above tools utilize empirical formulas to predict variance and control weights, they lack the capability to correct pseudorange biases. To address this need, we developed \textbf{pyrtklib}. This library, written in C++ and integrated into Python, combines the efficiency of C with the ease of use of Python.

\begin{table*}[]
    \centering
    \caption{Comparison of existing GNSS tools}
    \begin{tabular}{cccccc}
    \toprule
         Tool & Language & Open-source & Weight Prediction & Bias Correction & Weight+Bias Prediction\\
    \midrule
         RTKLIB & C & Y & Y & N & N\\
         Bernese GNSS & Fortran & N & Y & N & N\\
         MuSNAT & Matlab/C++ & N & Y & N & N\\
         NavSU & Matlab & Y & Y & N & N\\
         MAAST & Matlab & Y & Y & N & N\\
         goGPS & Matlab & Y & Y & N & N\\
         Laika & Python & Y & Y & N & N\\
         gnss\_lib\_py & Python & Y & Y & N & N\\
         \textbf{pyrtklib} & Python & Y & Y & Y & Y\\
    \bottomrule
    \end{tabular}
    \label{GNSS_tools}
\end{table*}

\subsection{Deep learning in GNSS}
In our recent review paper\cite{xu2024machine}, we categorize the application of deep learning in GNSS as follows:
\begin{enumerate}
    \item \textbf{Improved pseudorange measurement} \label{ipm}
    \item \textbf{Measurement status prediction} \label{msp}
    \item \textbf{Positioning level information} \label{pli}
    \item \textbf{Measurement error prediction} \label{mep}
\end{enumerate}

The approach outlined in \ref{ipm}) aims to enhance the correlator and discriminator at the receiver level to improve signal quality control\cite{borhani2023deep,li2022deep,orabi2020machine}. These methods are highly integrated, forming a super tightly coupled relationship between deep learning and GNSS. However, upgrading the receiver hardware is challenging to scale rapidly. The objective of \ref{msp}) is to utilize deep learning to classify NLOS and multipath signals in advance, thus eliminating or mitigating their adverse effects\cite{yakkati2022machine,guillard2023benefits,blais2022novel,zawislak2022gnss,jiang2022convolutional,guillard2023using,zhang2021prediction,suzuki2021nlos,munin2020convolutional,suzuki2020nlos,cho2019enhancing,quan2018convolutional,cho2023mpcnet,zhang2023learning}. The goal of \ref{pli}) is to predict the final positioning error and apply corrections\cite{mohanty2023learning,kanhere2022improving,tao2021real,siemuri2021improving,van2021end,xu2024framework}. These strategies represent preprocess and postprocess applications that are loosely coupled with deep learning in GNSS. Lastly, the algorithms discussed in \ref{mep}) have a direct impact on the measurements and are considered tightly coupled\cite{hu2023fisheye,zhang2021prediction,cho2023mpcnet,maaref2021leveraging}.

Super tightly coupled methods require support from either hardware or software-based receivers. In contrast, the labels for loosely coupled methods are easily accessible when the ground truth position is known, allowing the training process to be decoupled from the positioning process. However, for tightly coupled methods, measurement correction is intricately linked to positioning, making it impractical to separate training and positioning processes. In such scenarios, it is advantageous for both positioning and training to be conducted in the same programming language. To this end, our framework built on \textbf{pyrtklib} is specifically designed to support tightly coupled deep learning and GNSS approaches. We anticipate that our framework will facilitate the development of tightly coupled algorithms.

\section{Tightly Coupled Deep Learning \\ Framework For GNSS}

\subsection{GNSS position principle}

A standard GNSS model can be formulated as:

\begin{equation}
    p=r+c\Delta{t}+I+T+\epsilon
\end{equation}
\noindent Here, $p$ represents the pseudorange measured by the receiver, $c$ is the speed of light, and $I$ and $T$ signify the ionospheric and tropospheric delays, respectively. These delays are typically estimated using atmospheric models in single point positioning. $\Delta{t}$ denotes the receiver's time bias and $\epsilon$ represents Gaussian noise. The variable $r$ denotes the distance between the satellite and the receiver, which is formulated as follows:
\begin{equation}
    r = \sqrt{(x^s-x_r)^2+(y^s-y_r)^2+(z^s-z_r)^2}
\end{equation}
\noindent where $x^s$, $y^s$, $z^s$ and $x_r$, $y_r$, $z_r$ are the coordinates of the satellite and the receiver in Earth-centered, Earth-fixed (ECEF) coordinate system respectively. When there are $n$ pseudorange measurements, the observation function is:
\begin{equation}
        \widetilde{\mathbf{Z}} = h(\mathbf{y}) = \left[\begin{matrix}
        r_1+c\Delta{t}+I_1+T_1\\
        r_2+c\Delta{t}+I_2+T_2\\
        \ldots\\
        r_n+c\Delta{t}+I_n+T_n\\
    \end{matrix}\right]
\end{equation}
\noindent where $\widetilde{\mathbf{Z}}=[p_1,p_2\ldots,p_n]^T$ and $\mathbf{y}=(x_r,y_r,z_r,\Delta{t})$, which are the measurements and state respectively. The first-order approximation of the function can be written as:
\begin{equation}
    h(\mathbf{y}+\Delta{\mathbf{y}})\approx h(\mathbf{y})+\mathbf{H}\Delta\mathbf{y}
\end{equation}
\noindent $\mathbf{H}$ is the Jacobian matrix:
\begin{equation}
    \label{H_matrix}\mathbf{H}=\left[\begin{matrix}
        \frac{x^1-x_1}{r_1}&\frac{y^1-y_1}{r_1}&\frac{z^1-z_1}{r_1}&c\\
        \frac{x^2-x_1}{r_2}&\frac{y^2-y_1}{r_2}&\frac{z^2-z_1}{r_2}&c\\
        \ldots&\ldots&\ldots&\ldots\\
        \frac{x^n-x_1}{r_n}&\frac{y^n-y_1}{r_n}&\frac{z^n-z_1}{r_n}&c\\
    \end{matrix}\right]\\
\end{equation}

The Gussian-Newton-based non-linear weight least squares (WLS) is employed to solve the unknown state $\mathbf{y}$ by iteration as follows:

\begin{equation}
    \label{lsq}
    \Delta\mathbf{y}=\left(\mathbf{H}^\mathbf{T}\mathbf{W}\mathbf{H}\right)^{-1}\mathbf{H}^\mathbf{T}\mathbf{W}\left(\widetilde{\mathbf{Z}}-h(\mathbf{y}_i)\right)
\end{equation}
    
\begin{equation}
    \label{iteration}\mathbf{y}_{i+1}=\mathbf{y}_i+\Delta\mathbf{y}
\end{equation}

\noindent where $\mathbf{W}$ is the weight square matrix, and $\mathbf{y}_0$, is the initial guess of the state, typically set to set $(0,0,0,0)$. The iteration process is halted once $||\Delta{y}||$ falls below a predefined threshold, at which point the final $\mathbf{y}$ represents the determined position.

Note that the WLS positioning process requires two inputs, $\mathbf{W}$ and $\widetilde{\mathbf{Z}}$, We simplify the expression using the following equation:

\begin{equation}
    \label{wls_solve}
    \mathbf{y} = f_{WLS}(\mathbf{W},\widetilde{\mathbf{Z}})
\end{equation}

Although the description above represents an ideal scenario, various factors such as NLOS and multipath effects, imprecise ephemerides, or receiver errors can introduce biases. Consequently, the model can be reformulated as follows:

\begin{equation}
\label{unmodeled_p}
    p=r+c\Delta{t}+I+T+b+\epsilon
\end{equation}
\noindent where $b$ is the unmodeled bias. To achieve more accurate positioning results despite these biases, two strategies are employed. The first strategy involves directly correcting the pseudorange measurements, while the second strategy entails down-weighting the unhealthy measurements. In the following subsection, we will introduce a tightly coupled deep learning and GNSS framework designed for bias correction and weight prediction.

\subsection{Tightly coupled deep learning/GNSS framework}
As illustrated in equation (\ref{wls_solve}), achieving optimal positioning results relies on accurately predicting the weights $\mathbf{W}$ or obtaining improved measurements $\widetilde{\mathbf{Z}}$. By utilizing the ground truth position and employing the mean square error (MSE) as the loss function, we can optimize both the measurements and weights using the following equations:
\begin{gather}
    \label{loss_func}L(\mathbf{y}_{gt},\mathbf{y}) = \frac{1}{2}(\mathbf{y}_{gt}-\mathbf{y})^2\\
    \label{pLpZ}\frac{\partial{L}}{\partial{\widetilde{\mathbf{Z}}}} = \frac{\partial{L}}{\partial{\mathbf{y}}}\frac{\partial{\mathbf{y}}}{\partial{\widetilde{\mathbf{Z}}}}\\
    \label{pLpW}\frac{\partial{L}}{\partial{\mathbf{W}}} = \frac{\partial{L}}{\partial{\mathbf{y}}}\frac{\partial{\mathbf{y}}}{\partial{\mathbf{W}}}
    % \widetilde{\mathbf{Z}}_{n+1} = \widetilde{\mathbf{Z}}_n - \eta\frac{\partial{L}}{\partial{\widetilde{\mathbf{Z}}}}\\
    % \mathbf{W}_{n+1} = \mathbf{W}_n - \eta\frac{\partial{L}}{\partial{\mathbf{W}}}
\end{gather}
\noindent Equation (\ref{loss_func}) defines the loss function. Equations (\ref{pLpZ}) and (\ref{pLpW}) demonstrate the gradients of $\widetilde{\mathbf{Z}}$ and $\mathbf{W}$, respectively, calculated using the chain rule under the specified conditions of the loss function. Should $\mathbf{W}$ and $\widetilde{\mathbf{Z}}$ be derived from a neural network, these gradients are then propagated backward through the backpropagation process.

In this demonstration, three key features are selected as inputs:
\begin{itemize}
    \item \textbf{Carrier-to-noise density (C/N0)}: C/N0 is an essential parameter that quantifies the quality of the received signal. It is defined as the ratio of the carrier power to the noise power per unit bandwidth and is typically expressed in decibels-hertz (dB-Hz).
    \item \textbf{Elevation Angle}: The elevation angle is the vertical angle measured from the receiver's horizon to the line of sight of a satellite. This measurement indicates the satellite's position relative to the receiver's location on Earth. In urban environments, satellites with higher elevation angles are often line-of-sight (LOS) satellites and are less likely to be obstructed.
    \item \textbf{Residuals from Equal Weight Least Squares Solution}: Initially, the position is calculated using an equal weight least squares solution. The residuals from each measurement are then analyzed. This feature aids in identifying potentially problematic or unhealthy measurements, thereby enhancing the reliability of the positioning data.
\end{itemize}
These three features are compiled into a vector $\mathbf{x} = [C/N0, Elevation, Residual]$ for the network input.

\begin{figure}
    \centering
    \subfloat[]{\includegraphics[width=0.8\linewidth]{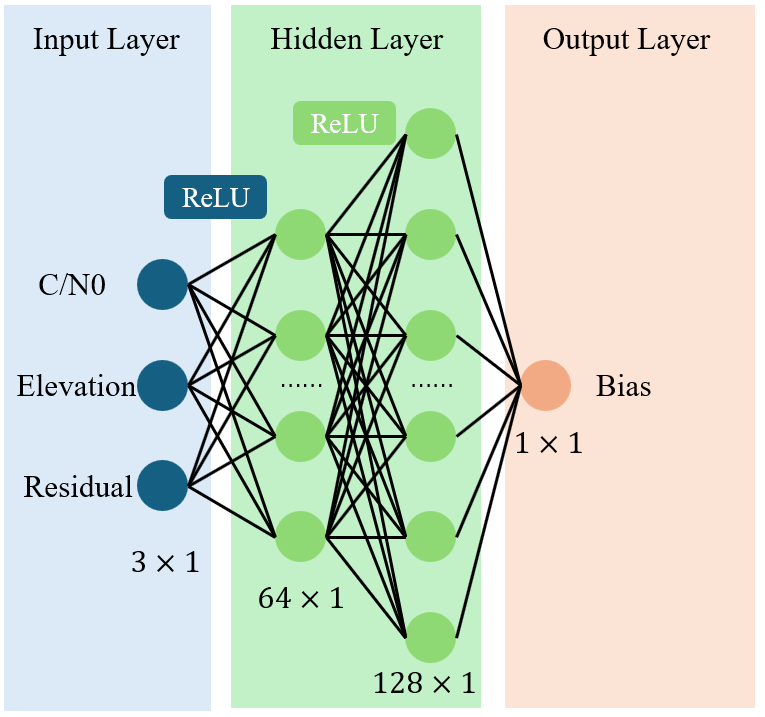}\label{biasnet}}\\
    \subfloat[]{\includegraphics[width=0.8\linewidth]{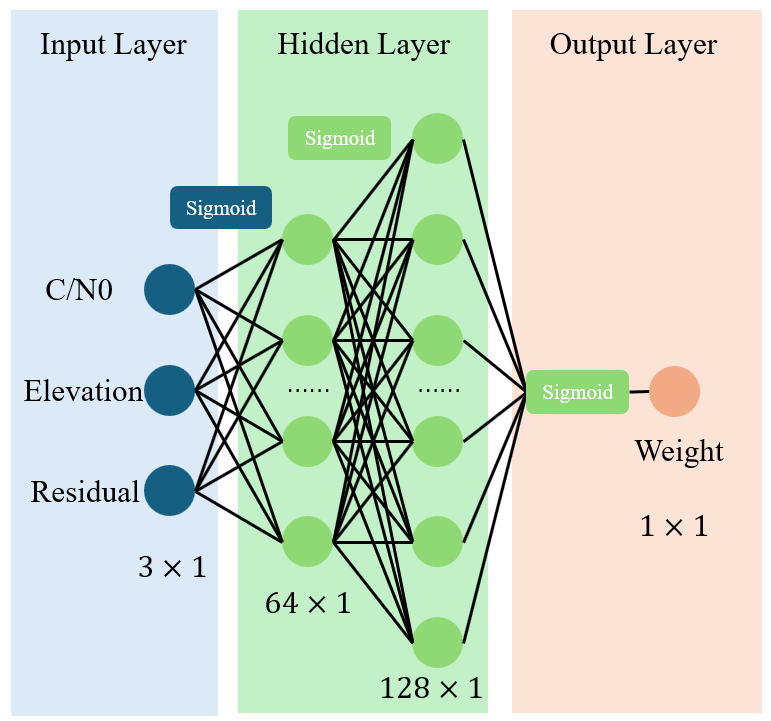}\label{weightnet}}\\
    \subfloat[]{\includegraphics[width=0.8\linewidth]{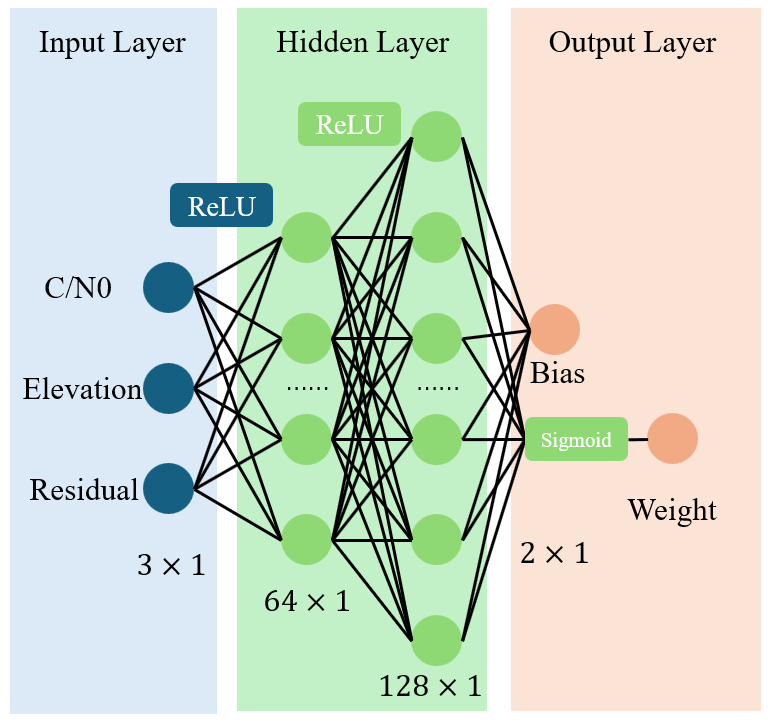}\label{biasweightnet}}
    \caption{The detailed structure of the bias network and weight network.}
    \label{network_structure}
\end{figure}

\subsubsection{Pseudorange bias correction network}
The detailed structure of the bias network is depicted in Figure \ref{biasnet}. This network comprises a straightforward four-layer architecture, including:
\begin{itemize}
    \item An input layer, which has a configuration of $3\times1$, corresponding to the three input features.
    \item Two hidden layers, sized $64\times1$ and $128\times1$ respectively, designed to progressively refine the feature representations.
    \item An output layer, configured as $1\times1$, which outputs the predicted bias.
\end{itemize}
The rectified linear unit (ReLU) is employed as the activation function throughout the network to introduce non-linearity, enhancing the model's capability to learn complex patterns. The final output represents the desired bias, which is used to predict the pseudorange bias as follows:
\begin{equation}
    \mathbf{b} = f_{nn,b}(\mathbf{X};\Theta)
\end{equation}
\noindent where $\mathbf{X}$ represents the batch input, defined as $\mathbf{X}=[\mathbf{x_1},\mathbf{x_2},\ldots,\mathbf{x_n}]$ and $\mathbf{b}=[b_1,b_2,\ldots,b_n]$ denotes the corresponding batch output. $\Theta$ symbolizes the set of parameters within the neural network. Using these definitions, the corrected measurements and resultant positions can be expressed as follows:
\begin{gather}
    \hat{\widetilde{\mathbf{Z}}} = \widetilde{\mathbf{Z}}-\mathbf{b}\\
    \hat{\mathbf{y}} = f_{WLS}(\mathbf{W},\hat{\widetilde{\mathbf{Z}}}) = f_{WLS}(\mathbf{W},\widetilde{\mathbf{Z}}-f_{nn,b}(\mathbf{X};\Theta))
\end{gather}
The training process can be formulated as follows:
\begin{gather}
    \label{trainB}\Theta= \underset{\Theta}{\text{argmin}}\sum_{i=1}^{n} L(\mathbf{y}_{gt} ,f_{WLS}(\mathbf{W},\widetilde{\mathbf{Z}}-f_{nn,b}(\mathbf{X}_i;\Theta)))\\
    \label{pLpbpth}\frac{\partial{L}}{\partial{\Theta}} = \frac{\partial{L}}{\partial{\hat{\mathbf{y}}}}\frac{\partial{\hat{\mathbf{y}}}}{\partial{\hat{\widetilde{\mathbf{Z}}}}}\frac{\partial{\hat{\widetilde{\mathbf{Z}}}}}{\partial\mathbf{b}}\frac{\partial\mathbf{b}}{\partial\Theta}\\
    \Theta_{n+1} = \Theta_n - \eta \frac{\partial{L}}{\partial{\Theta}}
\end{gather}
\noindent Equation (\ref{trainB}) illustrates that the training objective is to identify the optimal network parameters, $\Delta\Theta$, that minimize the loss function. Equation (\ref{pLpbpth}) details the gradient transfer process, with $\eta$ representing the learning rate.

\subsubsection{Weights prediction network}
The detailed structure of the weight network is depicted in Figure \ref{weightnet}. While similar to the bias network, this network employs a sigmoid activation function instead of ReLU. The network is designed to predict the weights as follows:
\begin{gather}
    \hat{\mathbf{w}} = f_{nn,w}(\mathbf{X};\Theta)\\
    \hat{\mathbf{W}} = diag(\mathbf{w})
\end{gather}
\noindent where $\hat{\mathbf{w}}=[w_1,w_2,\ldots,w_n]$ represents the vector of predicted weights in the batch output. $\hat{\mathbf{W}}$ is a diagonal matrix composed of the elements from $\hat{\mathbf{w}}$. The position process is then formulated as follows:

\begin{equation}
    \hat{\mathbf{y}} = f_{WLS}(\hat{\mathbf{W}},\widetilde{\mathbf{Z}}) = f_{WLS}(f_{nn,w}(\mathbf{X};\Theta),\widetilde{\mathbf{Z}})
\end{equation}

The training process is formulated as follows:
\begin{gather}
    \label{trainW}\Theta= \underset{\Theta}{\text{argmin}}\sum_{i=1}^{n} L(\mathbf{y}_{gt} ,f_{WLS}(f_{nn,w}(\mathbf{X}_i;\Theta),\widetilde{\mathbf{Z}}))\\
    \label{pLpwpth}\frac{\partial{L}}{\partial{\Theta}} = \frac{\partial{L}}{\partial{\hat{\mathbf{y}}}}\frac{\partial{\hat{\mathbf{y}}}}{\partial{\hat{\mathbf{W}}}}\frac{\hat{\mathbf{W}}}{\partial\Theta}\\
    \Theta_{n+1} = \Theta_n - \eta \frac{\partial{L}}{\partial{\Theta}}
\end{gather}
\noindent Equation (\ref{pLpwpth}) demonstrates the gradient transfer process within the tightly coupled deep learning and GNSS framework, tracing the path from the position loss function to the network parameters through the weights.

\subsubsection{Bias correction and weights prediction network}
The previously described frameworks focus exclusively on either bias or weight prediction, yet it is possible to predict both simultaneously. The architecture of this dual-prediction network is illustrated in Figure \ref{biasweightnet}. In this network, ReLU serves as the activation function for the initial two layers. The output layer features two outputs: one for bias and another for weight. To ensure that the weight values range from zero to one, a sigmoid function is applied specifically to the weight output. This design allows the bias and weight predictions to share network parameters, effectively extracting information from the input. The predicted bias and weight are denoted as follows:

\begin{gather}
    (\mathbf{b},\hat{\mathbf{W}}) = f_{nn,bw}(\mathbf{X};\Theta)\\
    \mathbf{b} = f_{nn,bw}^b(\mathbf{X};\Theta)\\
    \hat{\mathbf{W}} = f_{nn,bw}^w(\mathbf{X};\Theta)
\end{gather}
\noindent The superscript $b$ and $w$ represent the bias output and the weight output, respectively. With these outputs defined, the positioning process can be formulated as follows:

\begin{equation}
\begin{aligned}
    \hat{\mathbf{y}} &= f_{WLS}(\hat{\mathbf{W}},\widetilde{\mathbf{Z}}-\mathbf{b}) \\
                     &= f_{WLS}(f_{nn,bw}^w(\mathbf{X};\Theta),\widetilde{\mathbf{Z}}-f_{nn,bw}^b(\mathbf{X};\Theta))
\end{aligned}
\end{equation}

The training process is delineated as follows:
\begin{gather}
    \label{trainBW}\Theta= \underset{\Theta}{\text{argmin}}\sum_{i=1}^{n} L(\mathbf{y}_{gt} ,f_{WLS}(\hat{\mathbf{W}},\widetilde{\mathbf{Z}}-\mathbf{b}))\\
    \label{pLpbpwpth}\frac{\partial{L}}{\partial{\Theta}} = \frac{\partial{L}}{\partial{\mathbf{b}}}\frac{\partial{b}}{\partial{\Theta}}+\frac{\partial{L}}{\partial{\mathbf{W}}}\frac{\partial{W}}{\partial{\Theta}}\\
    \Theta_{n+1} = \Theta_n - \eta \frac{\partial{L}}{\partial{\Theta}}
\end{gather}
\noindent Equation (\ref{pLpbpwpth}) encapsulates how the gradients are derived from both the biases and weights, effectively linking the loss function to the network parameters $\Theta$ through tightly integrated feedback loops.

\section{Experiment}
In the preceding section, we detailed the training and prediction processes for our tightly coupled deep learning GNSS positioning framework. In this section, we will evaluate our approach and compare its performance against other tools. To facilitate a concise discussion, we will use the abbreviations TDL-B, TDL-W, and TDL-BW to refer to the bias correction network, weight prediction network, and combined bias correction and weight prediction network, respectively.

\subsection{Experiment Setup}

\begin{table}
    \centering
    \caption{The detail of the datasets.}
    \label{datasets}
    \begin{tabular}{cccccc}
        \toprule
         Dataset&Date&DoU&Epoch&Samples&Usage\\
         \midrule
         KLT1 &2021.06.10 & Light &203 & 4676 & testing\\
         KLT2 &2021.06.10 & Light&209 & 4914 & testing\\
         KLT3 &2021.06.10 & Light &404 & 8857 &training\\
         Whampoa &2021.07.14 & Deep & 1205 & 12926 &testing \\
         \bottomrule
    \end{tabular}
\end{table}

\begin{table}
    \caption{The detail of the sensors}
    \label{tab:device}
    \centering
    \begin{tabular}{cm{1cm}<{\centering}m{1.2cm}<{\centering}m{3.2cm}<{\centering}}
    \toprule
         Sensor & Output & Frequency(Hz) & Other \\
    \midrule
         SPAN-CPT & coordinate & 100 & / \\
         Ublox F9P & pseudorange & 1 & GPS L1, BeiDou B1, Galileo E1, GLONASS G1 \\
    \bottomrule
    \end{tabular}
    
\end{table}

Four datasets are utilized for evaluation. Three of these datasets were collected in the urban areas of Hong Kong's Kowloon Tong (KLT), and one dataset was gathered in the Whampoa area of Hong Kong. KLT is characterized as a light urban area, whereas Whampoa is considered a deep urban area with an approximate 2D positioning error of 18 meters. The specific details of each dataset are provided in Table \ref{datasets}, where DoU represents the degree of urbanization. KLT3 was designated for training, while the remaining datasets were used for testing.

A Ublox-F9P receiver was utilized to receive and decode GNSS signals at a frequency of 1Hz. Additionally, a NovAtel SPAN-CPT system\cite{kennedy2006architecture}, providing a Real-Time Kinematic (RTK) GNSS/INS integrated solution, was used to generate centimeter-level ground truth data at 100Hz. Further details are provided in Table \ref{tab:device}. These setups are consistent with those used in our previously open-sourced UrbanNav datasets\cite{hsu2021urbannav}.

\begin{table}
    \caption{Training details}
    \label{tab:training_details}
    \centering
    \begin{tabular}{m{2cm}<{\centering}m{6cm}<{\centering}}
    \toprule
        Component & Specification \\
    \midrule
         System & Ubuntu 20.04 \\
    \midrule
         CPU & AMD 5900x \\
    \midrule
         Graphics Card & Nvidia 4090 \\
    \midrule
         Memory & 128 Gigabytes \\
    \midrule
         Loss Function & Mean Square Error (MSE) \\
    \midrule
         Epoch & 500 for TDL-B and TDL-W, 100 for TDL-BW \\
    \midrule
         Optimizer & Adam \cite{kingma2014adam} \\
    \midrule
         Learing Rate & 0.001\\
    \bottomrule
    \end{tabular}
\end{table}

\begin{figure}
    \centering
    \includegraphics[width=1\linewidth]{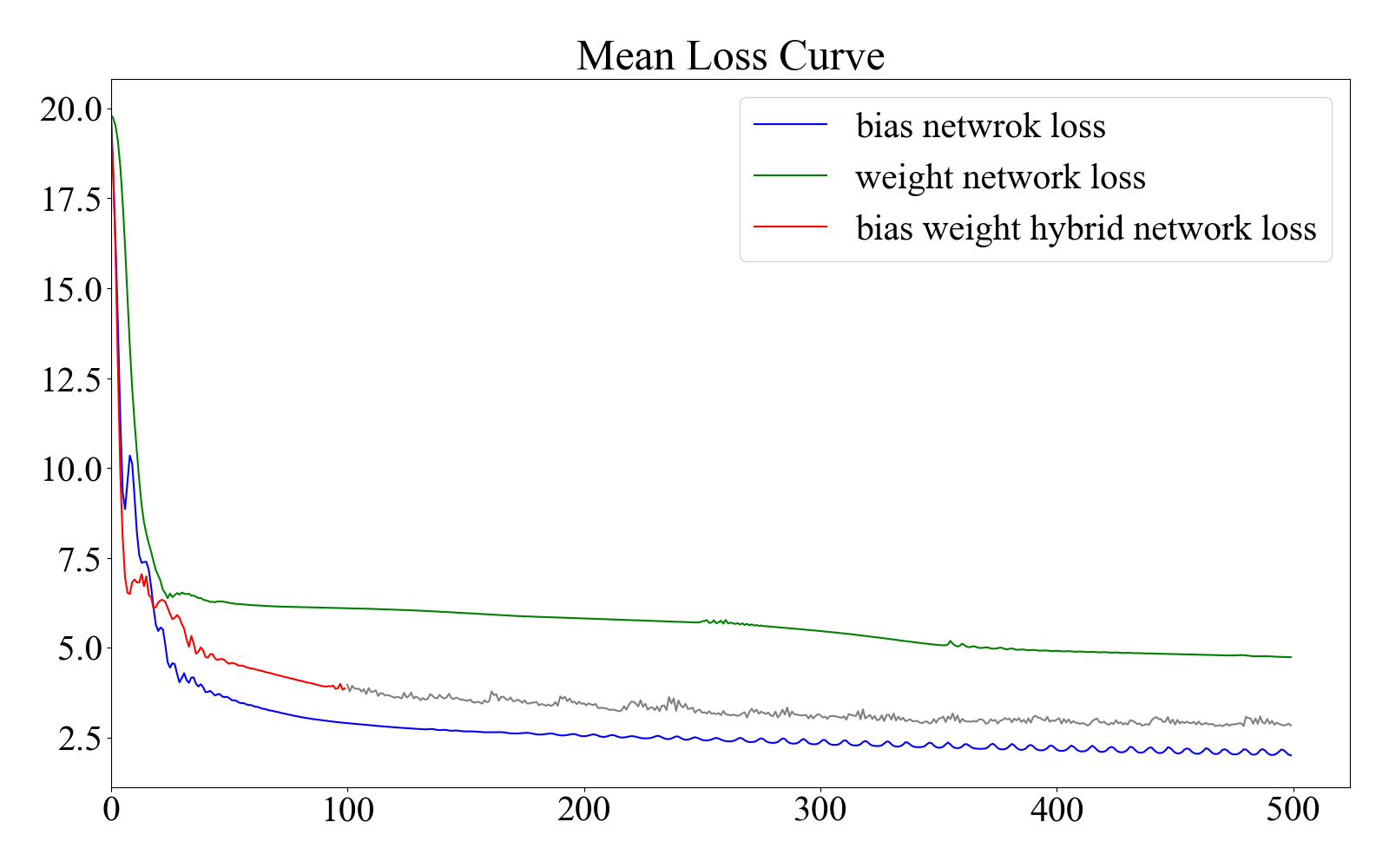}
    \caption{The loss curves of bias and weight network training process. The blue curve is the mean position loss of the bias network and the green curve is for weight network.}
    \label{fig:loss_curve}
\end{figure}

Details of the training process for the three networks are outlined in Table \ref{tab:training_details}. While most parameters are consistent across the networks, the number of training epochs varies. Specifically, the TDL-BW model utilizes fewer epochs to prevent overfitting, which is a risk due to its complex nature. The training loss curves for each network are illustrated in Figure \ref{fig:loss_curve}.

The training process involves a critical detail regarding the initial state $\mathbf{y}_0$, which should not be set as $(0,0,0,0)$. As depicted in Equations (\ref{trainW}), (\ref{trainB}), and (\ref{trainBW}), the process consists of a two-step optimization. Initially, the position state is solved using WLS optimization, and this solution is then tightly coupled to the network. The gradient of the weight in Equation (\ref{lsq}) largely depends on the $\mathbf{H}$ matrix, which is derived from the current position solution. Per Equations (\ref{lsq}) and (\ref{iteration}), the solution accumulates iteratively. In early iterations, significant changes in the solution result in drastic alterations to the $\mathbf{H}$ matrix, leading to unstable weight gradients. This instability can cause the optimization to fall into local minima and fail to converge. To mitigate this issue, the initial state is derived from the solution of an equal weight least squares calculation, which is nearly converged. Consequently, the $\mathbf{H}$ matrix remains relatively stable, enhancing the reliability of the training process.

\subsection{Experiment Result}
\subsubsection{Competing methods} In this section, we present our results and compare them with those obtained using RTKLIB and goGPS. Specifically, in RTKLIB, the weights assigned to each measurement are primarily derived from the elevation angles:
\begin{gather}
    \sigma^2(\theta)=a^2+\frac{b^2}{sin^2\theta}\\
    w = \frac{1}{\sigma^2(\theta)}
\end{gather}
\noindent In RTKLIB, the weight assigned to each measurement depends on the elevation angle, denoted by $\theta$. The coefficients $a$ and $b$, known as super parameters, are typically set at 0.3. Meanwhile, in goGPS\cite{herrera2016gogps}, weights are computed based on both the C/N0 and the elevation angle, as follows:

\begin{gather*}
    k_1(s) = -\frac{s-s_1}{a},k_2(s) = \frac{s-s_1}{s_0-s_1}
\end{gather*}

\begin{small}
\begin{equation}
    \begin{aligned}
        w=\begin{cases}
            \frac1{\sin^2\theta}\left(10^{k_1(S)}\left(\left(\frac A{10^{k_1(s_0)}}-1\right)k_2(S)+1\right)\right),S<s_1\\
            1, C/N0\geq s_1
        \end{cases}
    \end{aligned}
\end{equation}
\end{small}
\noindent In this formula, $S$ represents the C/N0, and $\theta$ denotes the elevation angle. The parameters $A$, $a$, $s_0$, and $s_1$ are super parameters and are typically set to 30, 20, 10, and 50, respectively. These values are crucial for determining the weights based on the quality and position of the satellite signals.

\begin{table}[]
    \centering
    \caption{2D mean error on testing datasets}
    \begin{tabular}{cccccc}%{m{0.8cm}<{\centering}m{1.6cm}<{\centering}m{1.9cm}<{\centering}m{1cm}<{\centering}m{1cm}<{\centering}}
    \toprule
    Dataset & \makecell{TDL-BW\\(m)}  & \makecell{TDL-B\\(m)}&\makecell{TDL-W\\(m)}&\makecell{goGPS\\(m)}&\makecell{RTKLIB\\(m)}\\
    \midrule
    KLT1    &  \textcolor{blue}{1.84} &  2.24       &    2.57       &    1.88       &    2.44\\
    KLT2    &  \textcolor{blue}{1.86} &  2.35       &    2.89       &    2.66       &     4.48\\
    Whampoa &  \textcolor{blue}{10.94} &  17.81      &    16.11       &    13.95       &     20.89\\
    \bottomrule
    \end{tabular}
    \label{tab:2d_error}
\end{table}

\begin{table}[]
    \centering
    \caption{3D mean error on testing datasets}
    \begin{tabular}{cccccc}%{m{0.8cm}<{\centering}m{1.6cm}<{\centering}m{1.9cm}<{\centering}m{1cm}<{\centering}m{1cm}<{\centering}}
    \toprule
    Dataset & \makecell{TDL-BW\\(m)}  & \makecell{TDL-B\\(m)}&\makecell{TDL-W\\(m)}&\makecell{goGPS\\(m)}&\makecell{RTKLIB\\(m)}\\
    \midrule
    KLT1    & \textcolor{blue}{4.72} &  5.30        &    9.92       &     18.14      &     10.31      \\
    KLT2    & \textcolor{blue}{3.92} &   5.89        &    7.75       &     15.44      &    10.94       \\
    Whampoa &  \textcolor{blue}{28.31} &  49.45        &    42.49       &     50.55      &    60.62       \\
    \bottomrule
    \end{tabular}
    \label{tab:3d_error}
\end{table}

\begin{figure}
    \centering
    \subfloat[]{\includegraphics[width=1\linewidth]{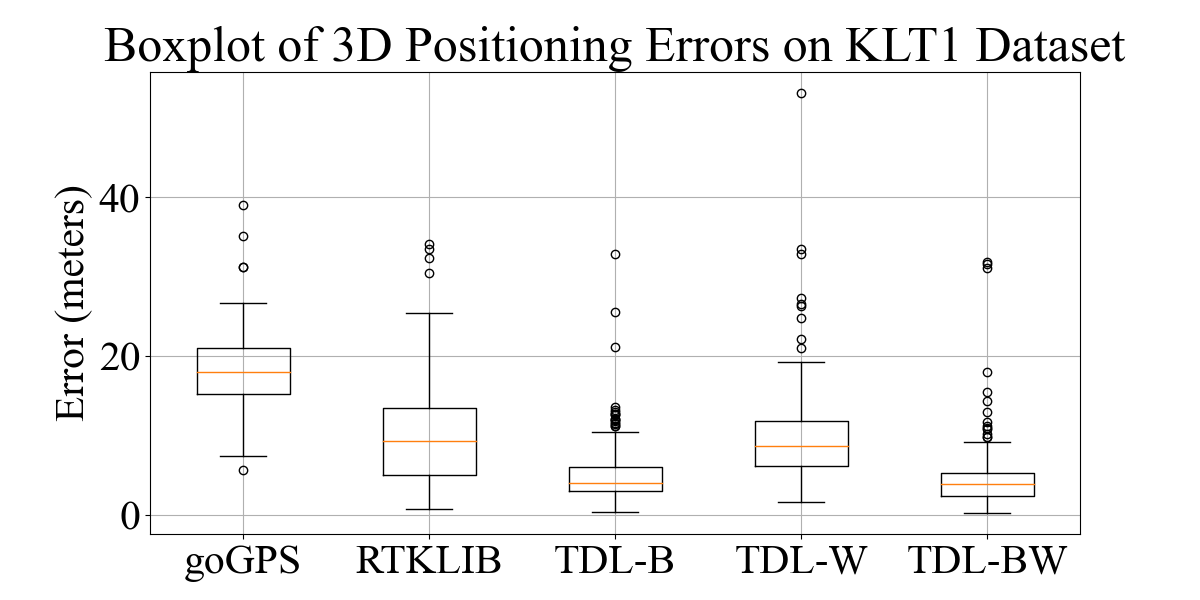}\label{klt1e3dbox}}\\
    \subfloat[]{\includegraphics[width=1\linewidth]{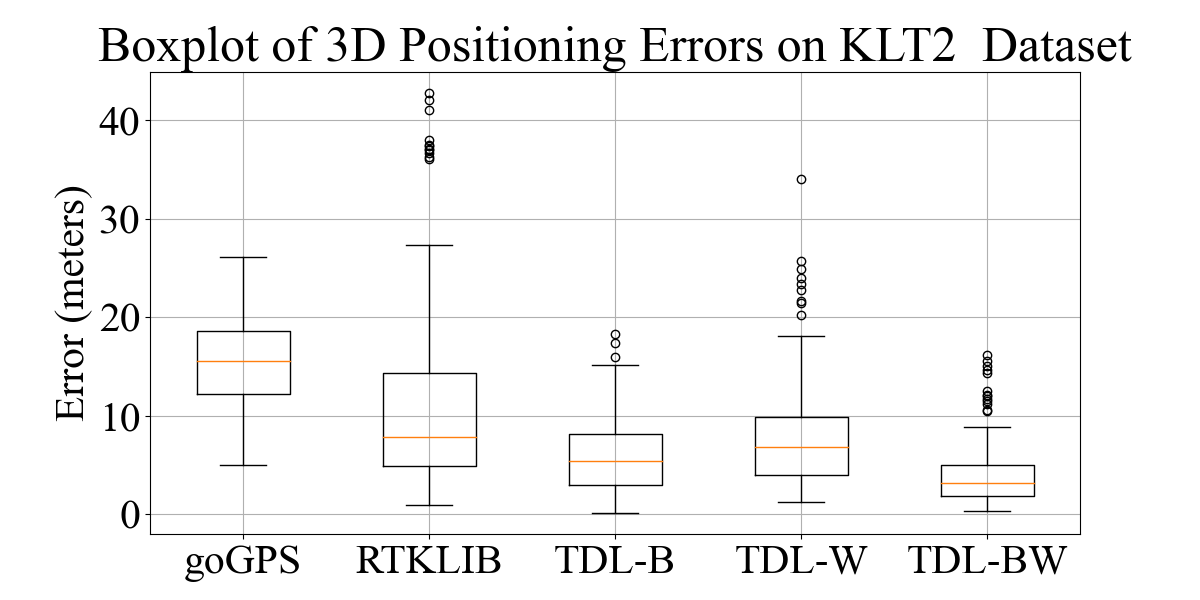}\label{klt2e3dbox}}\\
    \subfloat[]{\includegraphics[width=1\linewidth]{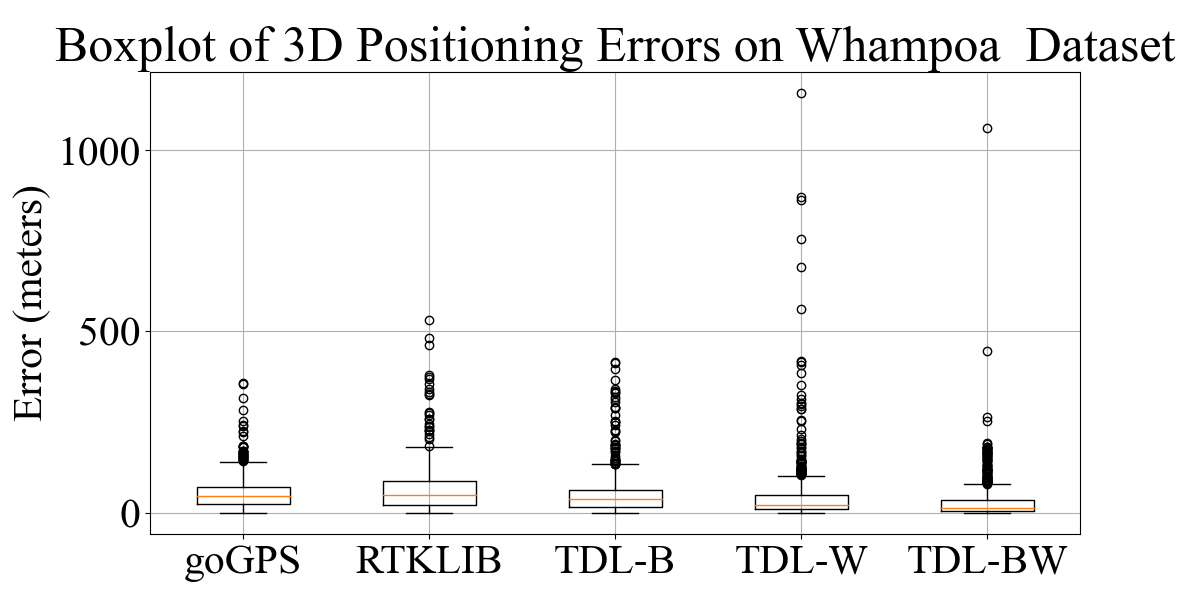}\label{we3dbox}}
    \caption{The boxplot for 3D error of the compared methods on the three datasets.}
    \label{fig_3derrorbox}
\end{figure}

\begin{figure}
    \centering
    \subfloat[]{\includegraphics[width=1\linewidth]{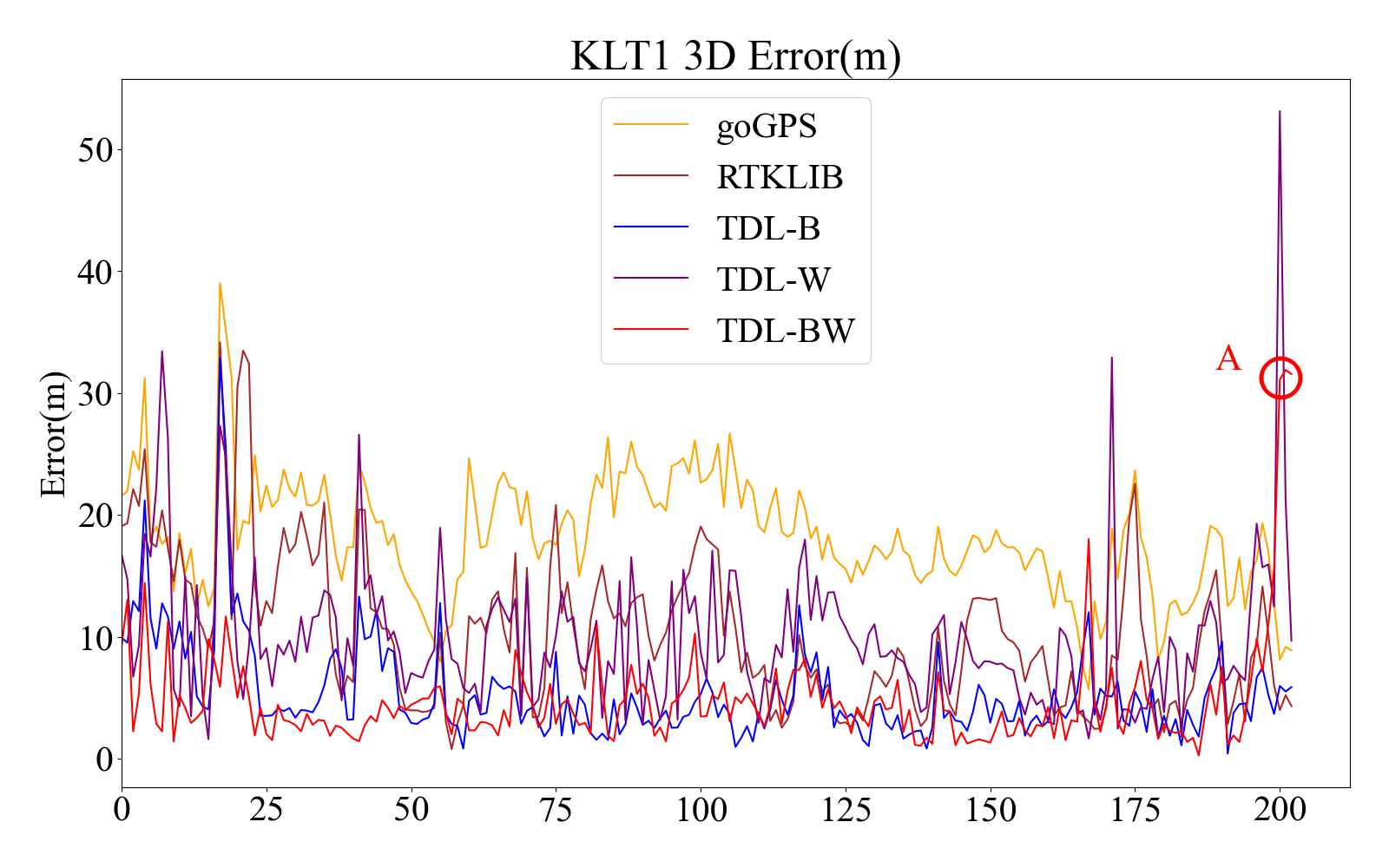}\label{klt1e3d}}\\
    \subfloat[]{\includegraphics[width=1\linewidth]{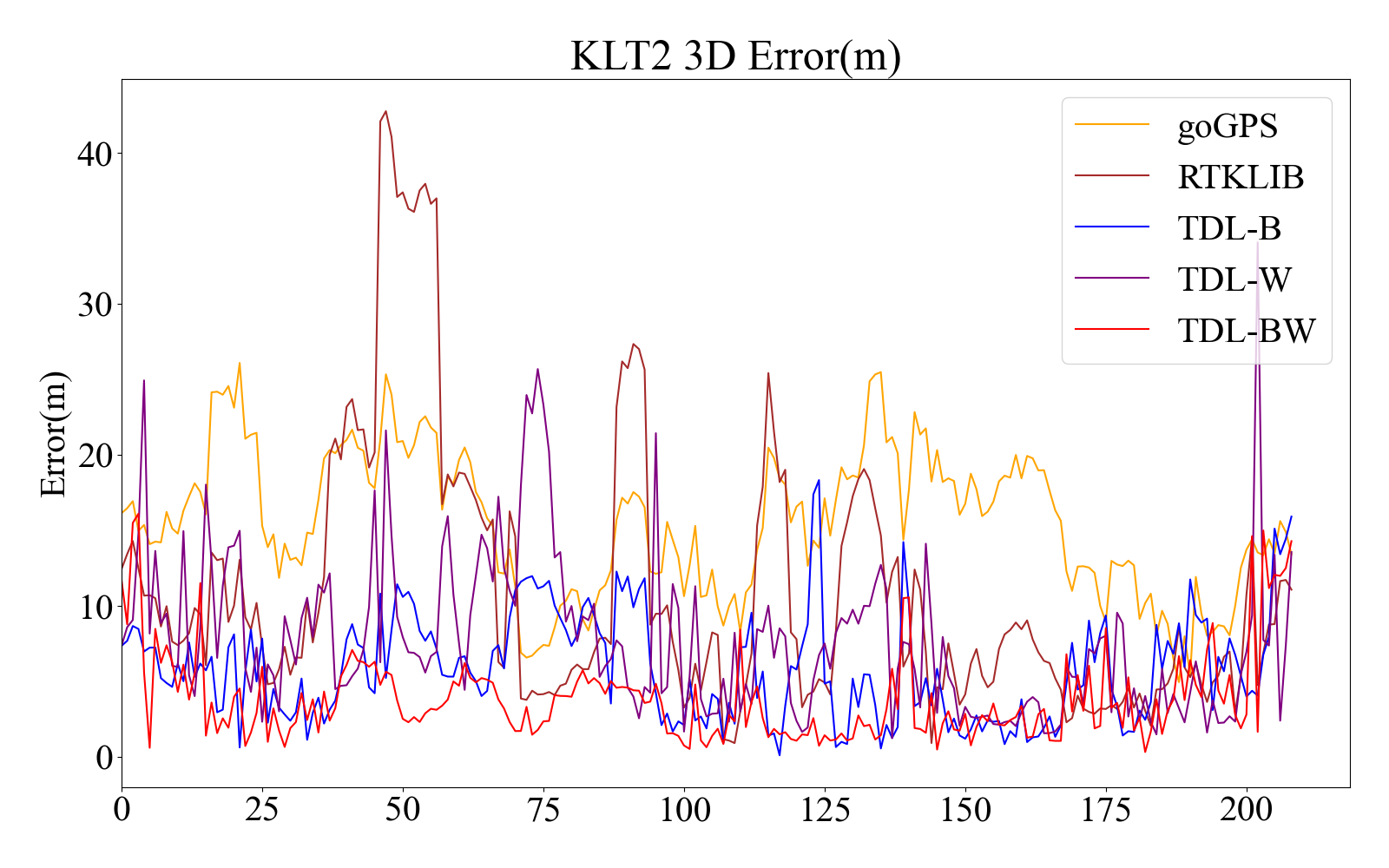}\label{klt2e3d}}\\
    \subfloat[]{\includegraphics[width=1\linewidth]{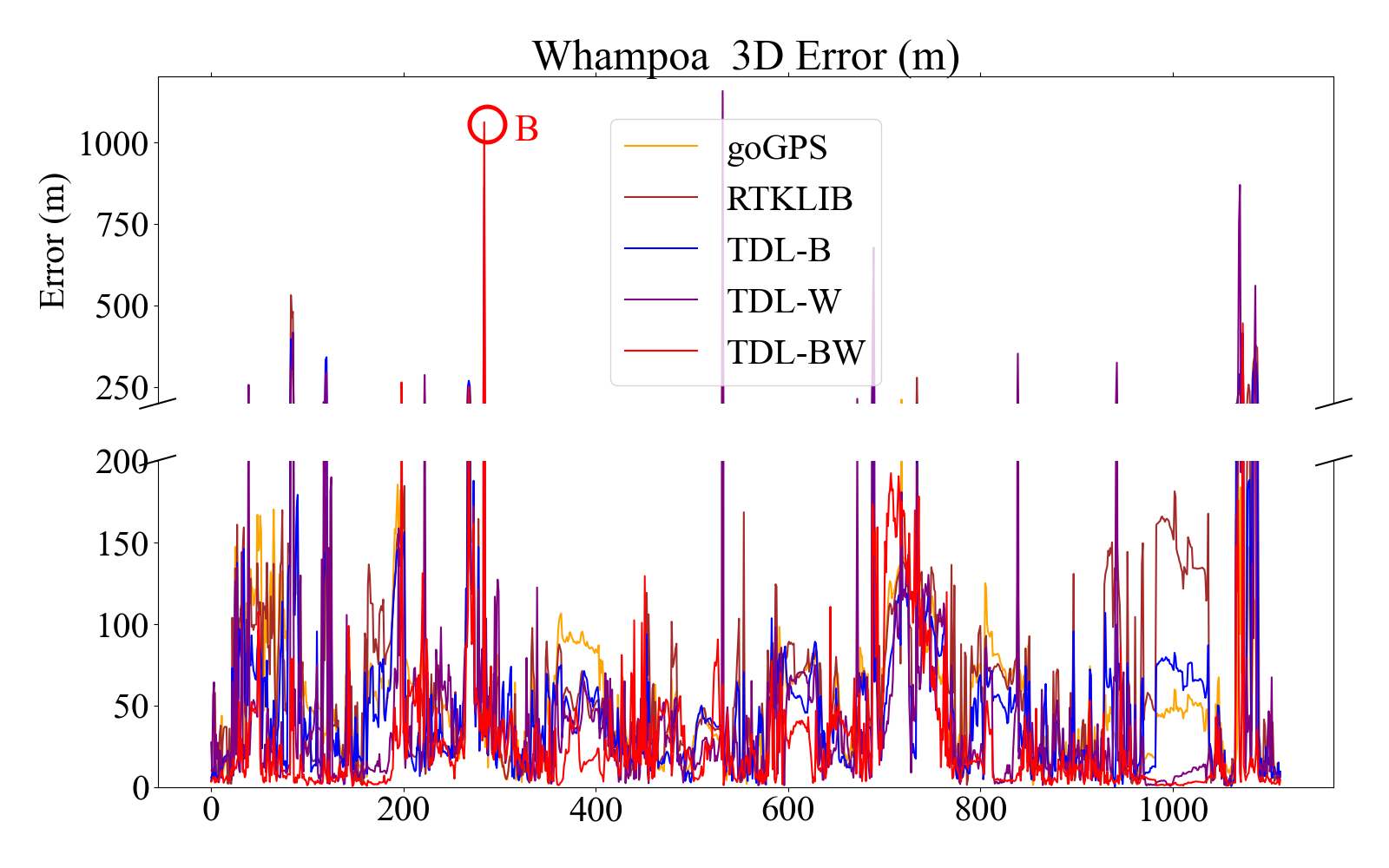}\label{we3d}}
    \caption{The detailed 3D error of the compared methods on the three datasets.}
    \label{fig_3derror}
\end{figure}

\begin{figure}
    \centering
    \subfloat[]{\includegraphics[width=1\linewidth]{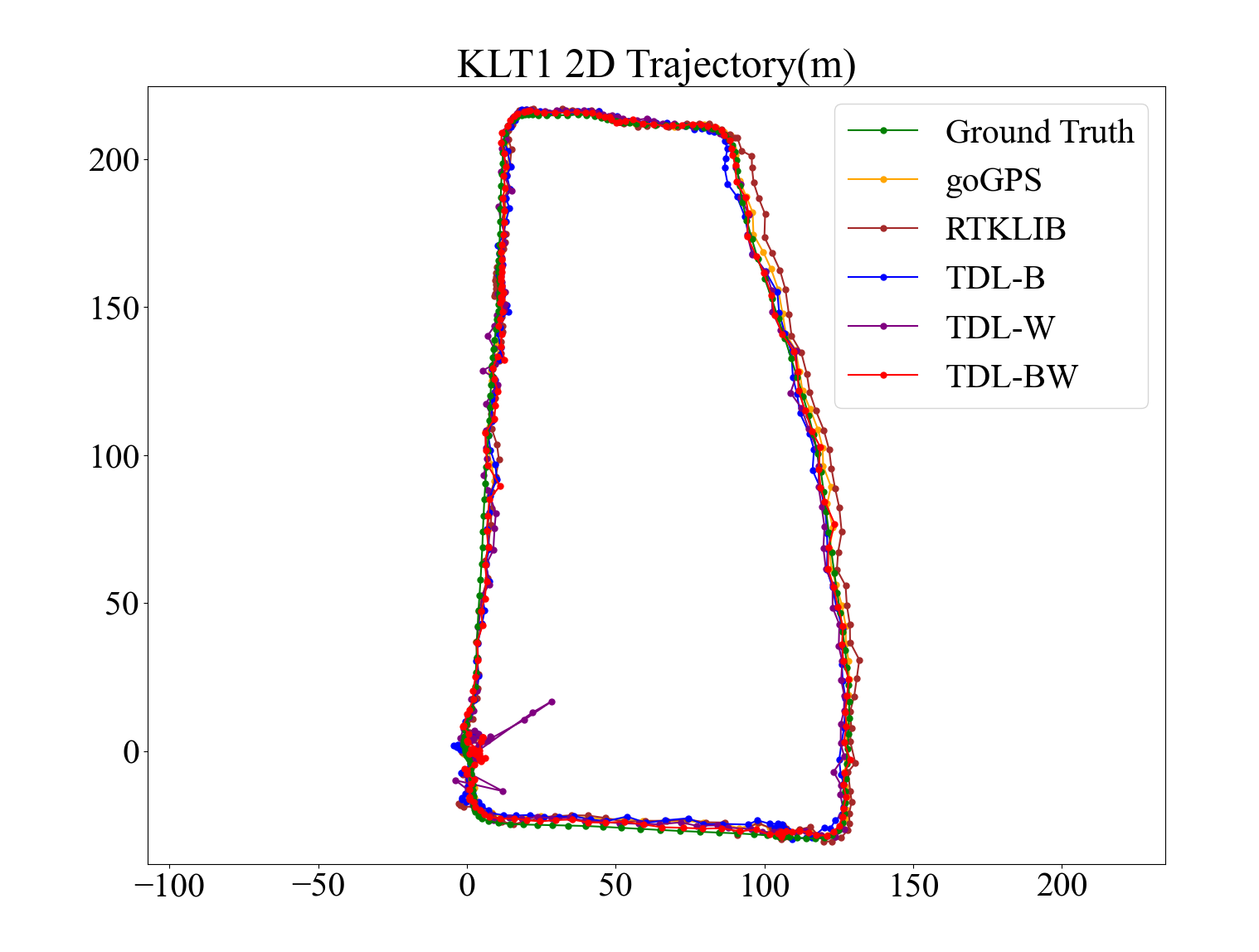}\label{klt1t2d}}\\
    \subfloat[]{\includegraphics[width=1\linewidth]{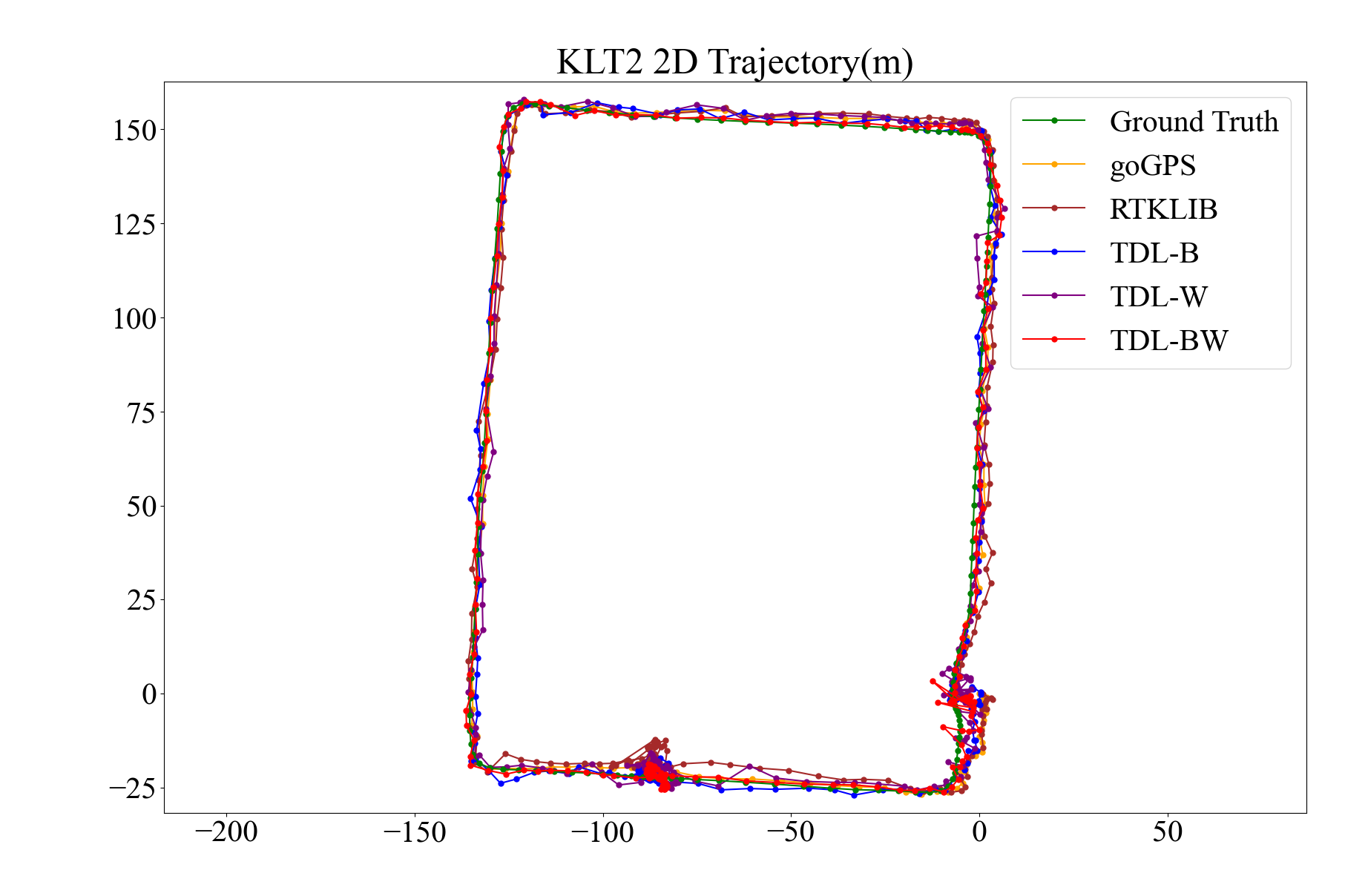}\label{klt2t2d}}\\
    \subfloat[]{\includegraphics[width=1\linewidth]{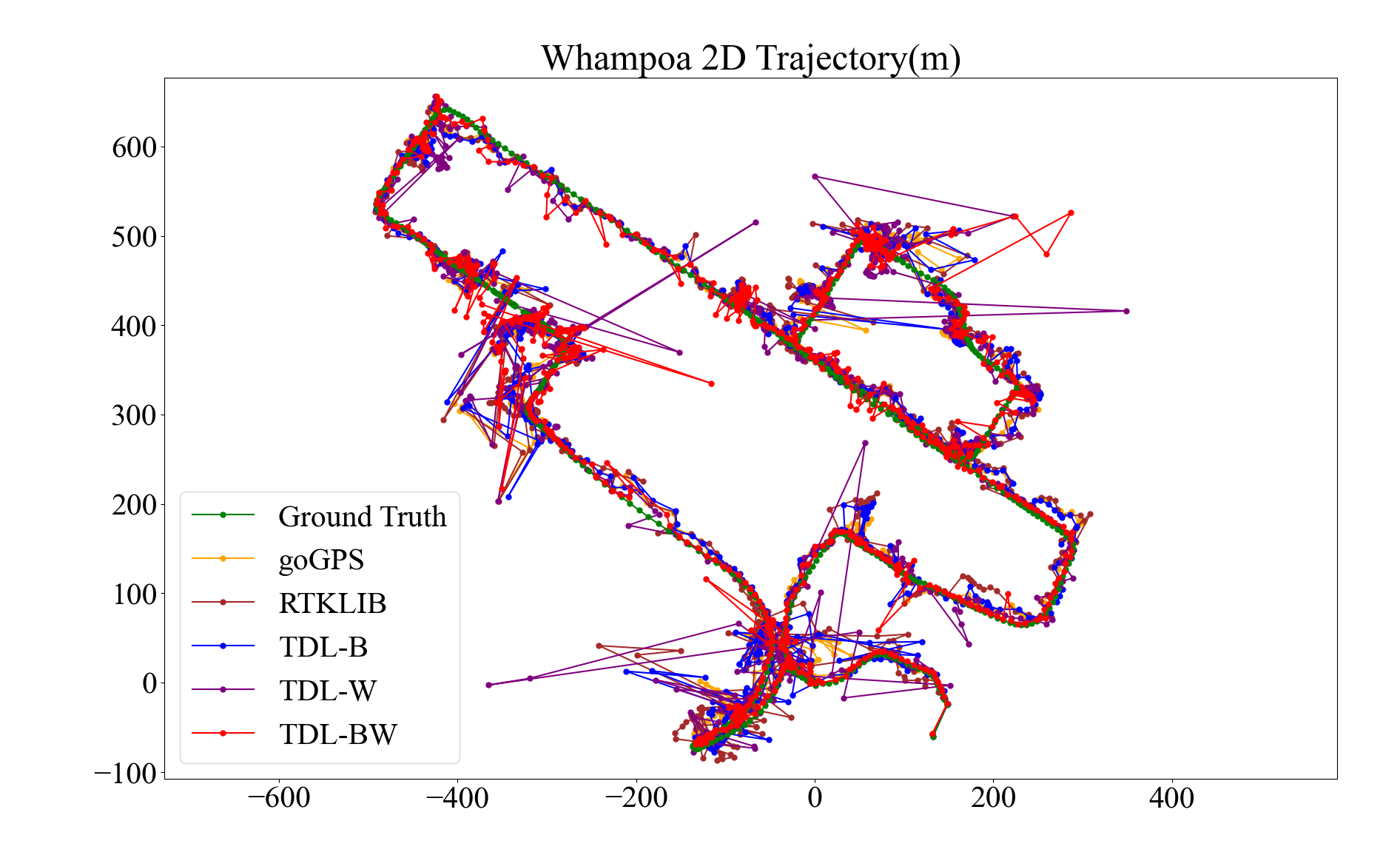}\label{wt2d}}
    \caption{The trajectory and ground truth of the compared methods on the three datasets.}
    \label{fig_traj}
\end{figure}

\subsubsection{Results and Analysis} The 2D and 3D positioning MSE errors are presented in Table \ref{tab:2d_error} and Table \ref{tab:3d_error}, respectively. Additionally, a boxplot of the 3D errors and a detailed view of these errors are depicted in Figure \ref{fig_3derrorbox} and Figure \ref{fig_3derror}. The corresponding trajectories are illustrated in Figure \ref{fig_traj}. In the 2D error comparison, the TDL-B and TDL-W models perform worse than goGPS; however, their 3D error results are significantly better than those of both goGPS and RTKLIB. This discrepancy arises because the training process primarily utilizes the 3D error to calculate the loss function, leading to a focused analysis on 3D errors.

According to Table \ref{tab:3d_error} and boxplot Figure \ref{fig_3derrorbox}, it is evident that the TDL-BW consistently outperforms others in terms of both accuracy and reliability. Specifically, TDL-BW exhibits the lowest mean errors and standard deviations, indicating a high level of precision and stability, which is crucial for applications requiring rigorous spatial accuracy. For instance, within the KLT1 dataset, TDL-BW achieved a mean error of 4.72 meters with a standard deviation of 4.26 meters, signifying minimal deviation in error measurement across samples. Similarly, in the KLT2 dataset, it maintained a mean error of 3.92 meters and an even lower standard deviation of 2.99 meters, further affirming its superior performance. Conversely, the weight strategies used by goGPS and RTKLIB demonstrated significant variability, particularly in the Whampoa dataset where deep urban challenging conditions likely exacerbated their performance issues; goGPS and RTKLIB recorded mean errors of 54.42 meters and 64.58 meters, respectively, coupled with high standard deviations exceeding 40 meters. These results highlight the variability and potential limitations of goGPS and RTKLIB in such conditions, suggesting a more suitable application in less demanding scenarios.

\section{Discussion}

\begin{table}[]
    \centering
    \caption{The predicted weight and bias in case A}
    \label{wb_caseA}
    \begin{tabular}{ccc|ccc}
    \toprule
         PRN & weight & bias & PRN & weight & bias\\
    \midrule
            \textcolor{green}{G07}&0.62&4.61&\textcolor{red}{G03}&0.00&5.44\\
            \textcolor{green}{G01}&0.36&3.12&\textcolor{red}{G14}&0.00&3.22\\
            \textcolor{red}{G30}&0.14&3.48&\textcolor{red}{G17}&0.00&7.81\\
            \textcolor{green}{C11}&0.12&2.27&\textcolor{green}{G22}&0.00&5.77\\
            \textcolor{green}{G21}&0.01&5.88&\textcolor{red}{C13}&0.00&6.00\\
            \textcolor{red}{C07}&0.00&4.72&\textcolor{red}{C08}&0.00&2.25\\
            \textcolor{red}{C23}&0.00&6.83&\textcolor{red}{C25}&0.00&2.19\\
            \textcolor{red}{G28}&0.00&4.41&&&\\
    \bottomrule
    \end{tabular}
\end{table}

\begin{table}[]
    \centering
    \caption{The predicted weight and bias in case B}
    \label{wb_caseB}
    \begin{tabular}{ccc|ccc}
    \toprule
         PRN & weight & bias & PRN & weight & bias\\
    \midrule
         \textcolor{green}{C10} & 1.00 & 0.40 & \textcolor{red}{G12} & 0.00 & 9.38 \\
         \textcolor{green}{G19} & 0.96 & 1.90 & \textcolor{red}{G14} & 0.00 & 8.32 \\
         \textcolor{red}{G06} & 0.89 & 3.46 & \textcolor{red}{G20} & 0.00 & 13.99 \\
         \textcolor{red}{G17} & 0.65 & 1.86 & \textcolor{red}{C08} & 0.00 & 2.32 \\
         \textcolor{green}{C07} & 0.21 & 0.22 & \textcolor{red}{C13} & 0.00 & 6.19 \\
         \textcolor{red}{G02} & 0.00 & 14.72 & \textcolor{red}{C28} & 0.00 & 20.50 \\
         \textcolor{red}{G05} & 0.00 & 11.85 & \textcolor{red}{C30} & 0.00 & 7.43 \\
    \bottomrule
    \end{tabular}
\end{table}

\begin{figure}
    \centering
    \subfloat[]{\includegraphics[width=1\linewidth]{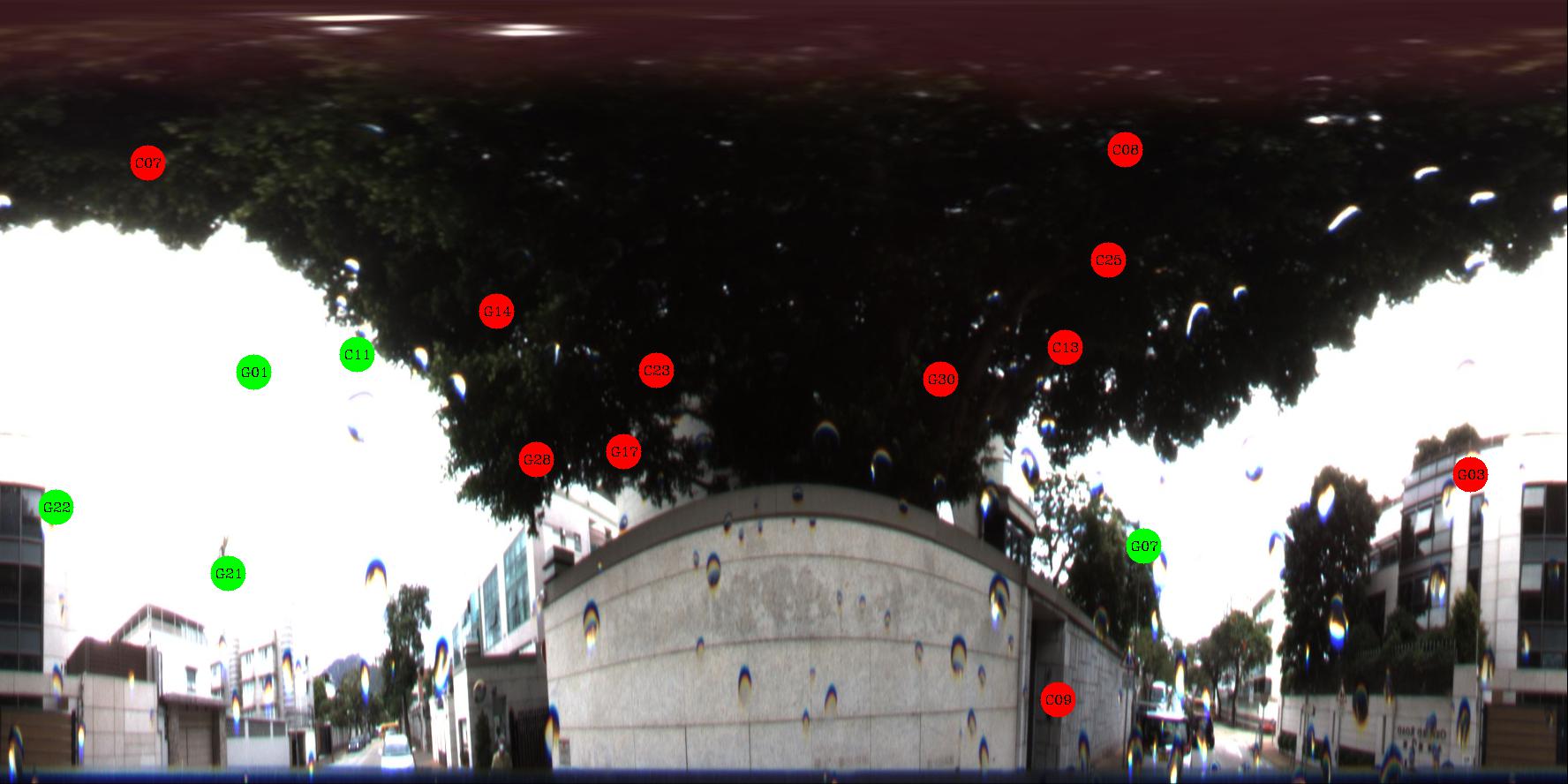}\label{figcaseA}}\\
    \subfloat[]{\includegraphics[width=1\linewidth]{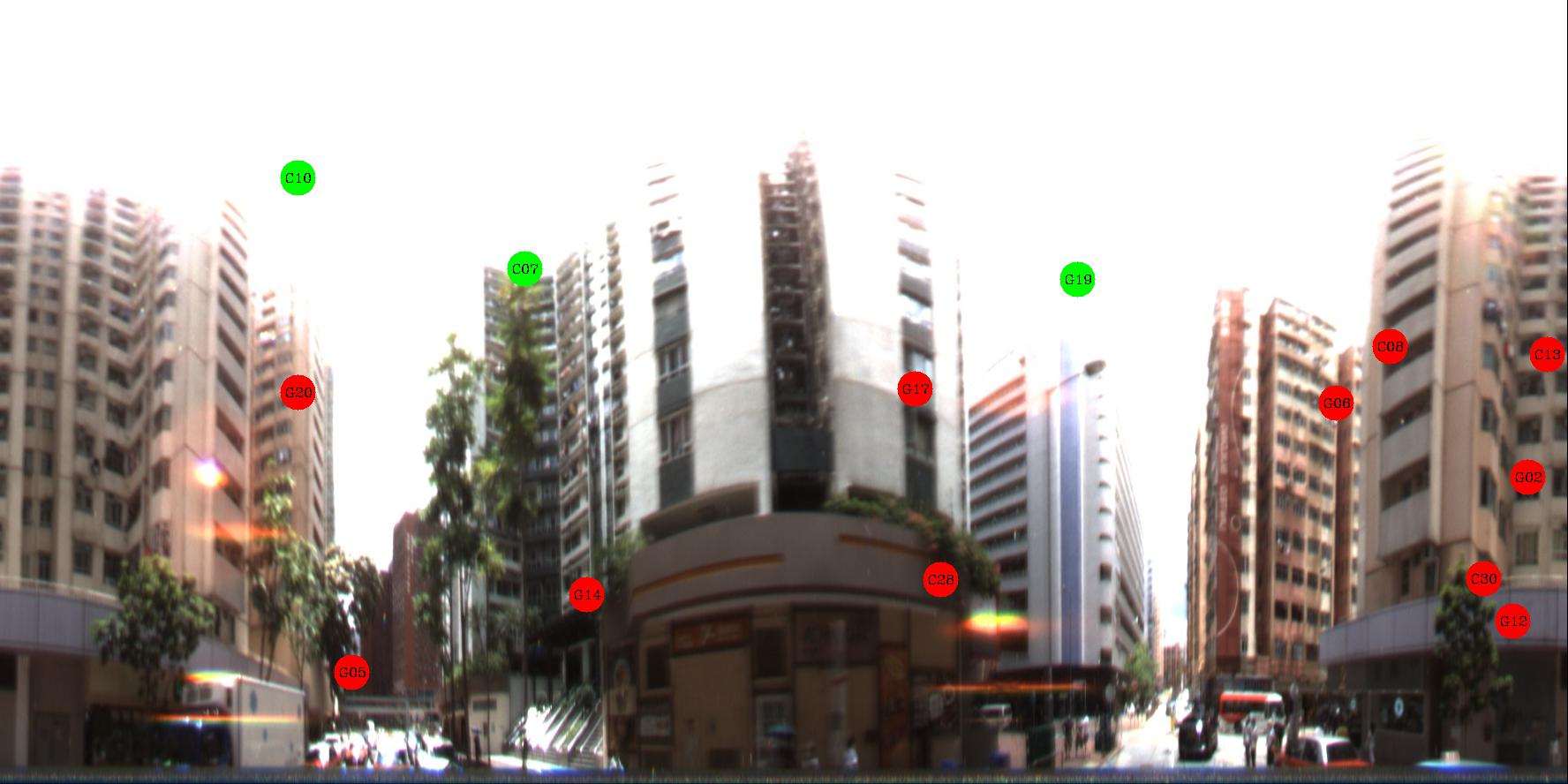}\label{figcaseB}}\\
    \caption{The scenarios of case A and B}
    \label{case_scene}
\end{figure}

In the section, we focus on two typical outliers of TDL-BW in the two representative scenarios, Case A and Case B, depicted in Figure \ref{case_scene}. The errors in the two cases are 31.89 meters and 1061.09 meters respectively. These scenarios highlight the challenges posed by LOS and NLOS conditions on satellite signal reception and the consequent effects on positioning accuracy.

In case A, as illustrated in Figure \ref{figcaseA} and detailed in Table \ref{wb_caseA}, the TDL-BW network outputs show a clear differentiation between LOS and NLOS signals, as indicated by the color-coded PRN entries (green for LOS and red for NLOS). This scenario, characterized by a tree canopy, predominantly exhibits NLOS conditions, impacting the weight distribution among satellites. Notably, satellites labeled as LOS (G07 and G01) receive non-zero weights, affirming the network's capability to identify viable signals amidst obstructions. However, the network misclassifies satellite G22, positioned near the edge of a building possibly affected by diffraction, assigning it no weight. A critical observation here is the excessive elimination of NLOS satellites, which, while reducing noise from obstructed signals, also minimizes redundancy in the available data for accurate positioning, potentially compromising the robustness of the positioning solution.

Presented in Figure \ref{figcaseB} and Table \ref{wb_caseB} for case B, this scenario demonstrates a similar pattern where the network effectively identifies and assigns higher weights to LOS satellites (C10 and G19). It also appropriately categorizes C07, which, despite potential obstructions, is deemed reliable. However, akin to Case A, the network's stringent filtering leads to the exclusion of numerous NLOS satellites, manifesting in an overly sparse dataset that might detract from the accuracy of the resultant positioning due to insufficient satellite coverage and geometry.

Both cases underscore the TDL-BW network's proficiency in distinguishing between LOS and NLOS satellites and its consequential decision-making concerning weight assignments. While this ability is advantageous for enhancing signal quality by excluding NLOS influences, it also raises concerns regarding the adequacy of satellite data for reliable positioning. The elimination of too many satellites, particularly under dense canopy or urban settings where NLOS conditions are prevalent, could severely limit the system's operational effectiveness by reducing the geometric diversity necessary for optimal positioning. The less robust performance of TDL-W in several situations is also due to the failure to allocate weights to enough measurements. In contrast, the results from TDL-B in the two cases are 5.52 meters and 223.50 meters, which are much better than TDL-BW and TDL-W. Therefore, due to the unreliable results from TDL-W and TDL-BW, where only a few measurements are weighted, it is advisable to switch to TDL-B.

\section{Conclusion}

\begin{figure}
    \centering
    \includegraphics[width=1\linewidth]{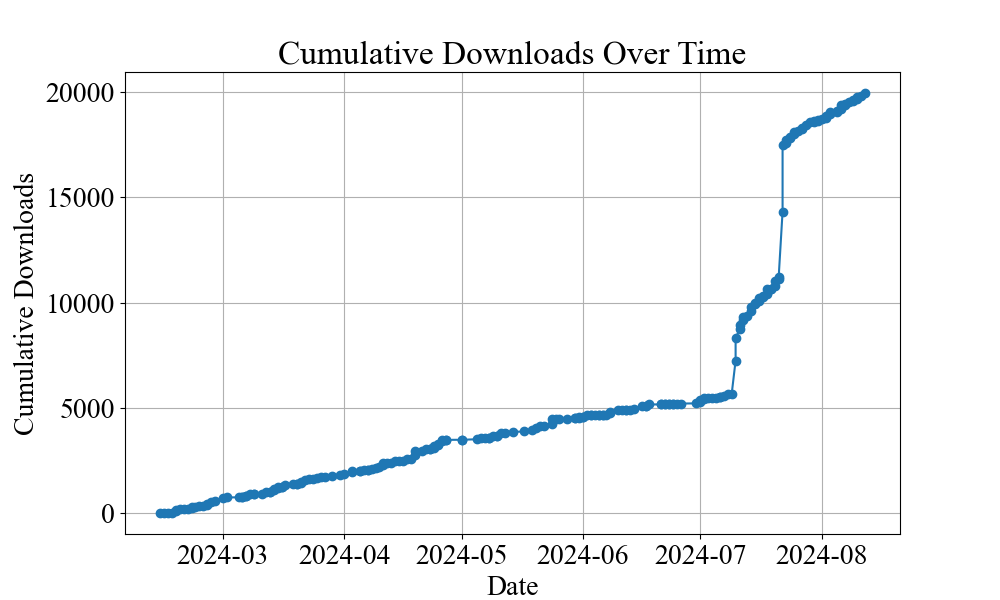}
    \caption{The cumulative download number for \textbf{pyrtklib} over the past 6 months}
    \label{download_curve}
\end{figure}

In this paper, we introduced \textbf{pyrtklib}, a Python binding for the widely-used GNSS library, RTKLIB. Utilizing \textbf{pyrtklib}, we developed a tightly coupled deep learning subsystem that predicts weights and biases for each satellite, thereby enhancing positioning performance. Our methods were compared against RTKLIB and goGPS. The results demonstrate that TDL-BW, which simultaneously predicts both weights and biases, outperforms the others. This network effectively differentiates between line-of-sight (LOS) and non-line-of-sight (NLOS) satellites, assigning appropriate weights and biases accordingly. Both \textbf{pyrtklib} and the deep learning subsystem are available as open-source resources at \href{https://github.com/IPNL-POLYU/pyrtklib}{https://github.com/IPNL-POLYU/pyrtklib} and \href{https://github.com/ebhrz/TDL-GNSS}{https://github.com/ebhrz/TDL-GNSS}. As shown in Figure \ref{download_curve}, \textbf{pyrtklib} has been downloaded approximately 20,000 times over the past six months.

The network structure employed in this demonstration is relatively straightforward, and the feature set used is limited. Looking ahead, our framework is designed to seamlessly incorporate a broader range of deep learning approaches. We plan to enhance the network architecture to account for spatial and temporal variations, and to integrate multi-modal inputs such as images, point clouds, and maps. Our aim is to contribute significantly to the community by bridging the gap between AI and GNSS technologies, enriching the potential applications and effectiveness of both fields.

\section*{Acknowledgments}
This project is sponsored by the Research Centre for Data Sciences \& Artificial Intelligence (RCDSAI). We also wish to express our gratitude for the support provided by the Meituan Academy Of Robotics Shenzhen (H-ZGHQ). And we sincerely thank our colleagues Ivan Ng, Liyuan Zhang and Dr. Yingying Wang.

% {\appendix[Proof of the Zonklar Equations]
% Use $\backslash${\tt{appendix}} if you have a single appendix:
% Do not use $\backslash${\tt{section}} anymore after $\backslash${\tt{appendix}}, only $\backslash${\tt{section*}}.
% If you have multiple appendixes use $\backslash${\tt{appendices}} then use $\backslash${\tt{section}} to start each appendix.
% You must declare a $\backslash${\tt{section}} before using any $\backslash${\tt{subsection}} or using $\backslash${\tt{label}} ($\backslash${\tt{appendices}} by itself
%  starts a section numbered zero.)}


% \section{References Section}
% You can use a bibliography generated by BibTeX as a .bbl file.
%  BibTeX documentation can be easily obtained at:
%  http://mirror.ctan.org/biblio/bibtex/contrib/doc/
%  The IEEEtran BibTeX style support page is:
%  http://www.michaelshell.org/tex/ieeetran/bibtex/
 
\bibliographystyle{IEEEtran}

\bibliography{ref}

\newpage

\section{Biography Section}
\vspace{-35pt}
\begin{IEEEbiography}[{\includegraphics[width=1in,height=1.25in,clip,keepaspectratio]{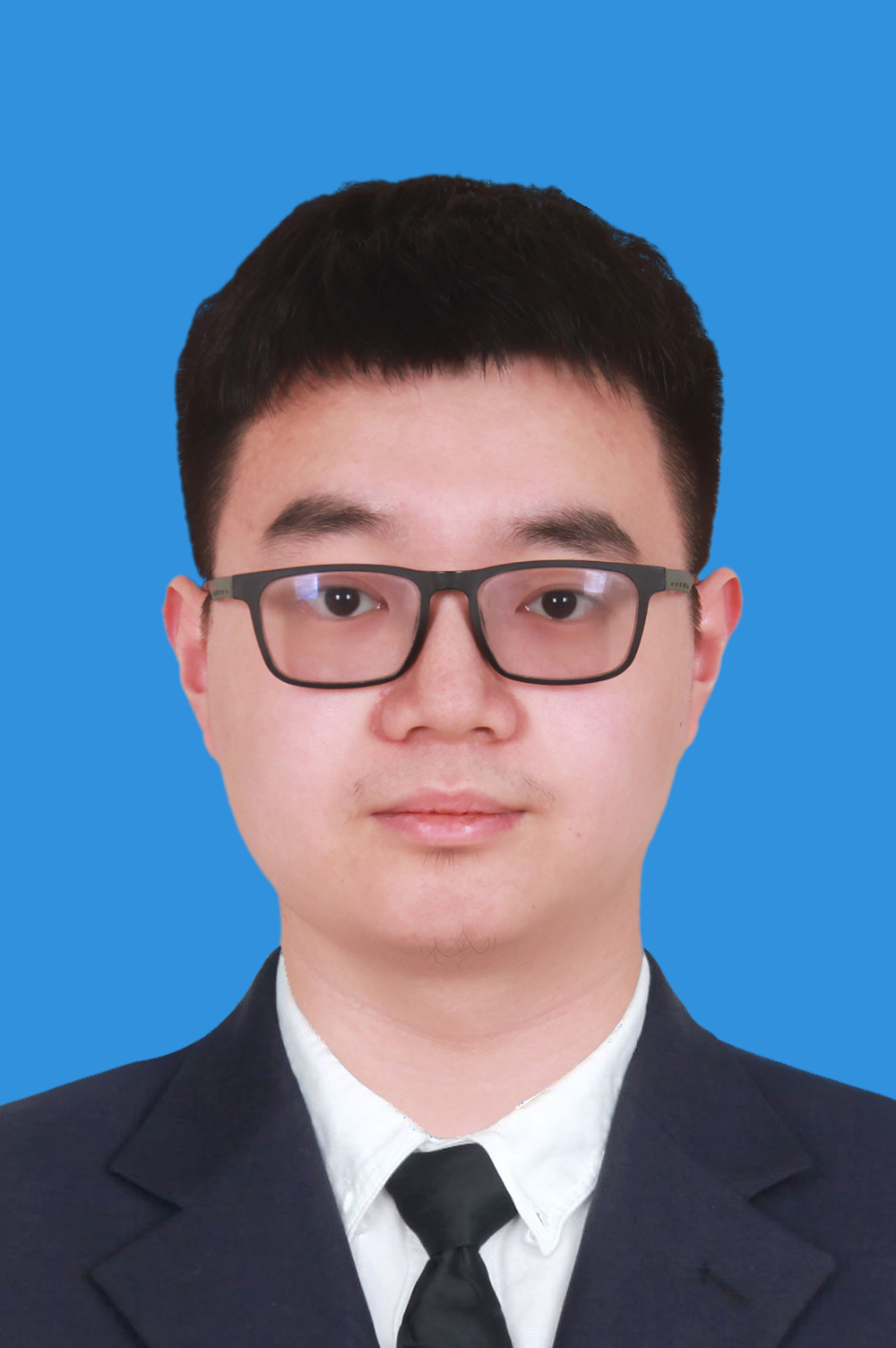}}]{Runzhi Hu}
was born in Leshan, Sichuan, China. He received his B.S and master degrees in mechanical engineering and computer science, respectively, from China Agricultural University. He now is a Ph.D candidate at the Hong Kong Polytechnic University. His research interests include HD map, multi-sensor fusion, SLAM, and GNSS positioning in urban canyons.
\end{IEEEbiography}

\begin{IEEEbiography}[{\includegraphics[width=1in,height=1.25in,clip,keepaspectratio]{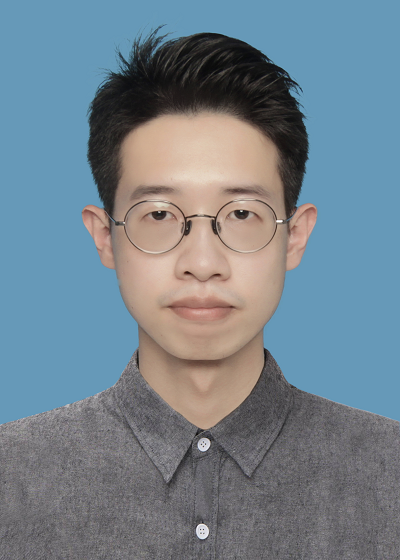}}]{Penghui Xu}
(Graduate Student Member, IEEE) received a B.S. degree from South China Agricultural University in 2015. In 2017, he obtained his MSc degree in mechanical engineering from The Hong Kong Polytechnic University. After that, he mainly works in machine learning algorithm development. Currently, he is a Ph.D. candidate at The Hong Kong Polytechnic University. His research interests include machine learning, GNSS urban localization, and multi-sensor integration for positioning.
\end{IEEEbiography}

\begin{IEEEbiography}[{\includegraphics[width=1in,height=1.25in,clip,keepaspectratio]{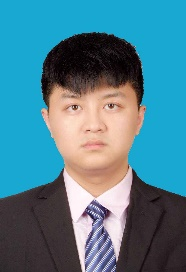}}]{Yihan Zhong}
received the bachelor’s degree in process equipment and control engineering from Guangxi University, Nanning, China, in 2020, and the master’s degree from The Hong Kong Polytechnic University (PolyU), Hong Kong, in 2022, where he is currently pursuing the Ph.D. degree with the Department of Aeronautical and Aviation Engineering (AAE). His research interests include collaborative positioning and low-cost localization.
\end{IEEEbiography}

\begin{IEEEbiography}[{\includegraphics[width=1in,height=1.25in,clip,keepaspectratio]{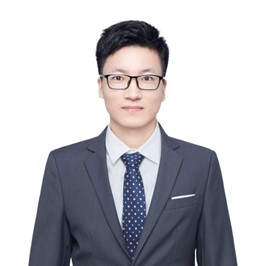}}]{Weisong Wen}
(Member, IEEE) received a BEng degree in Mechanical Engineering from Beijing Information Science and Technology University (BISTU), Beijing, China, in 2015, and an MEng degree in Mechanical Engineering from the China Agricultural University, in 2017. After that, he received a PhD degree in Mechanical Engineering from The Hong Kong Polytechnic University (PolyU), in 2020. He was also a visiting PhD student with the Faculty of Engineering, University of California, Berkeley (UC Berkeley) in 2018. Before joining PolyU as an Assistant Professor in 2023, he was a Research Assistant Professor at AAE of PolyU since 2021. He has published 30 SCI papers and 40 conference papers in the field of GNSS (ION GNSS+) and navigation for Robotic systems (IEEE ICRA, IEEE ITSC), such as autonomous driving vehicles. He won the innovation award from TechConnect 2021, the Best Presentation Award from the Institute of Navigation (ION) in 2020, and the First Prize in Hong Kong Section in Qianhai-Guangdong-Macao Youth Innovation and Entrepreneurship Competition in 2019 based on his research achievements in 3D LiDAR aided GNSS positioning for robotics navigation in urban canyons. The developed 3D LiDAR-aided GNSS positioning method has been reported by top magazines such as Inside GNSS and has attracted industry recognition with remarkable knowledge transfer.
\end{IEEEbiography}


\vfill

\end{document}